\documentclass[runningheads]{llncs}
\usepackage[T1]{fontenc}
\usepackage{amsmath}
\usepackage{amssymb}
\usepackage{graphicx}
\usepackage{wrapfig}
\usepackage{hyperref}
\usepackage{color}

\begin{document}
\title{MagBotSim: Physics-Based Simulation and Reinforcement Learning Environments for Magnetic Robotics}
\titlerunning{MagBotSim: Simulation and RL Environments for Magnetic Robotics}
%
\author{Lara Bergmann\inst{1}\orcidID{0009-0005-0544-2291} \and
Cedric Grothues\inst{1}\orcidID{0009-0004-3367-6661} \and
Klaus Neumann\inst{1,2}\orcidID{0009-0008-7221-0441}}
\authorrunning{L. Bergmann et al.}
%

\institute{CITEC, Bielefeld University, Inspiration 1, 33619 Bielefeld, Germany \\
\email{\{lara.bergmann,cedric.grothues,klaus.neumann\}@uni-bielefeld.de}
\and
Fraunhofer IOSB-INA, Campusallee 1, 32657 Lemgo, Germany}
\maketitle              

\begin{abstract}
Magnetic levitation is about to revolutionize in-machine material flow in industrial automation. Such systems are flexibly configurable and can include a large number of independently actuated shuttles (movers) that dynamically rebalance production capacity. Beyond their capabilities for dynamic transportation, these systems possess the inherent yet unexploited potential to perform manipulation. By merging the fields of transportation and manipulation into a coordinated swarm of magnetic robots (MagBots), we enable manufacturing systems to achieve significantly higher efficiency, adaptability, and compactness. To support the development of intelligent algorithms for magnetic levitation systems, we introduce \emph{MagBotSim} (\emph{Magnetic Robotics} Simulation): a physics-based simulation for magnetic levitation systems. By framing magnetic levitation systems as robot swarms and providing a dedicated simulation, this work lays the foundation for next generation manufacturing systems powered by \emph{Magnetic Robotics}. 

MagBotSim's documentation, videos, experiments, and code are available at: \url{https://ubi-coro.github.io/MagBotSim/}
\end{abstract}
\section{Introduction}
\begin{wrapfigure}{r}{0.29\textwidth}
  \centering
  \vspace{-0.8cm}
  \includegraphics[width=0.29\textwidth]{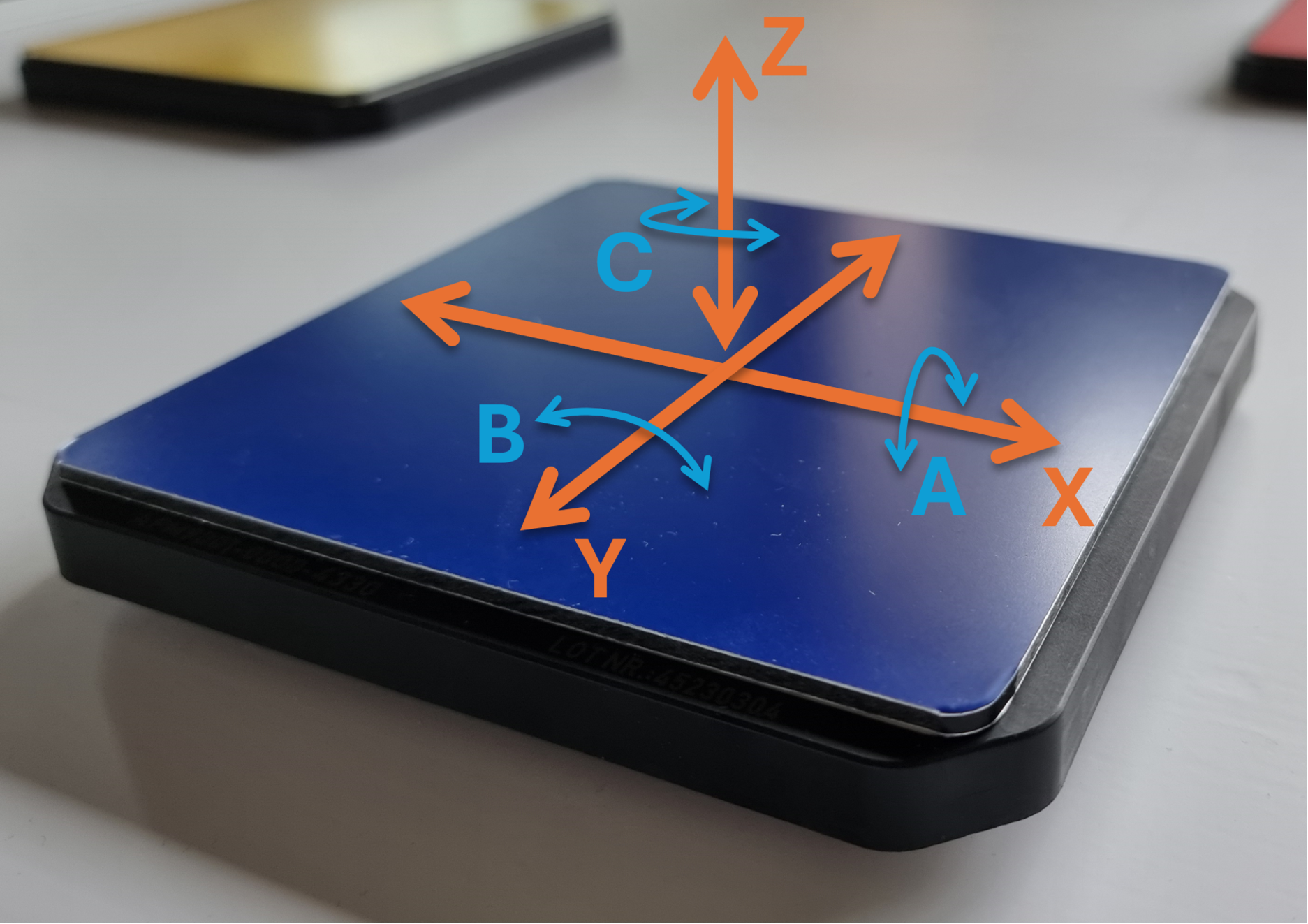}
  \caption{Mover on a magnetic levitation system}
  \label{fig_mover_dofs}
  \vspace{-0.7cm}
\end{wrapfigure}
\label{sec_intro}
\begin{figure}[h!]
  \centering
  \includegraphics[width=0.8\linewidth]{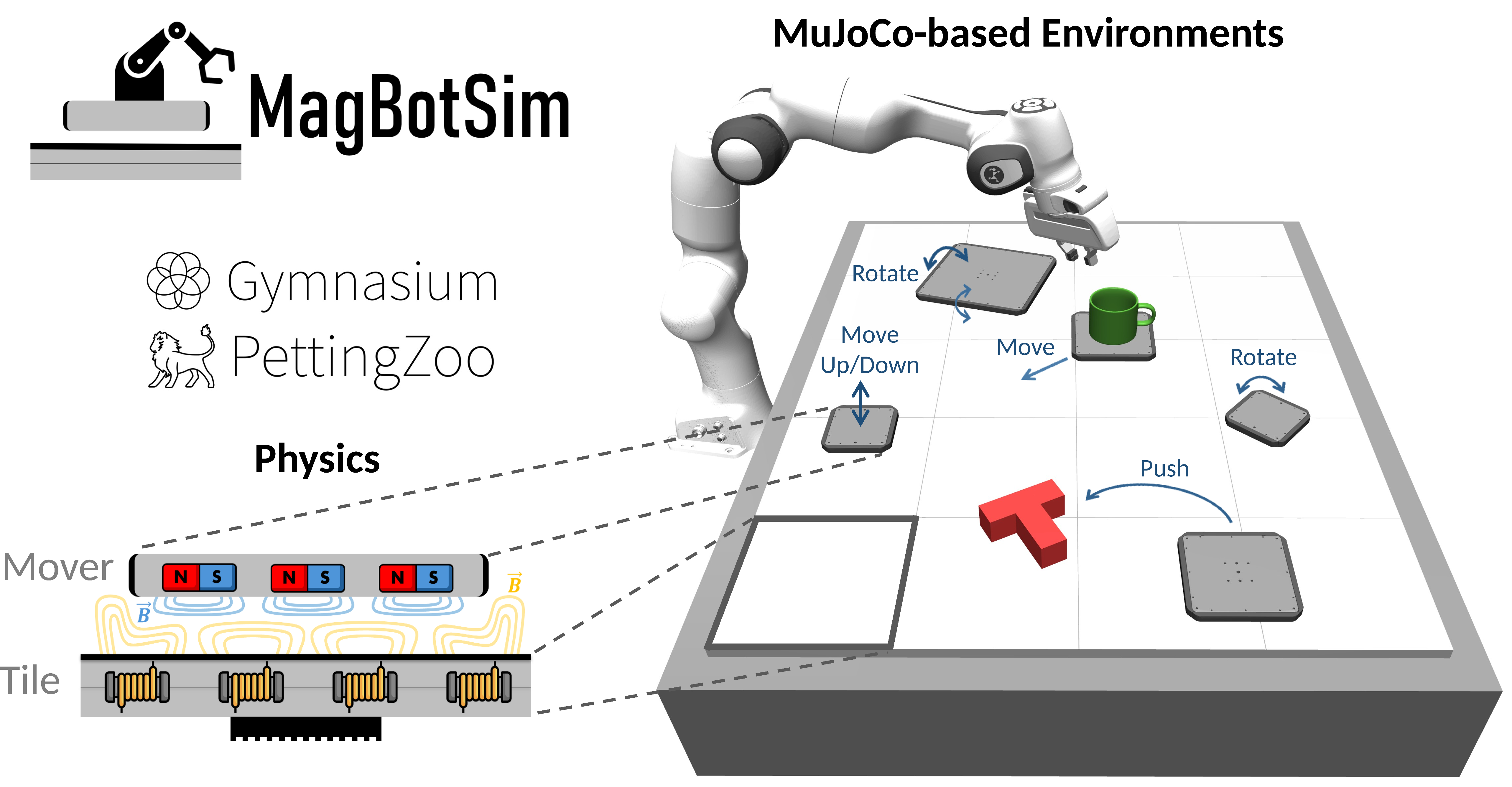}
  \caption{Schematic overview of the proposed MagBotSim library, a physics-based simulation for \emph{Magnetic Robotics}. MagBotSim includes reinforcement learning environments for trajectory planning and object manipulation.}
  \label{fig_visual_abstact}
\end{figure} 
Electric drive technologies based on magnetic levitation are about to revolutionize inline-machine material flow and individualized product transport in manufacturing systems due to additional degrees of freedom (DoF), high configurability, and the resulting flexibility. These so-called \emph{magnetic levitation systems} (MagLev systems) consist of two basic components, as shown in Fig.~\ref{fig_visual_abstact}. 
Firstly, dynamically actuated shuttles as passive motor modules, so-called \emph{movers} (see Fig.~\ref{fig_mover_dofs}), consist of a housing and a complex permanent magnet structure based on Halbach arrays~\cite{lu_6d_2012} on the lower side of the mover. Secondly, static motor modules, so-called \emph{tiles}, are the active component of the drive system. As shown in Fig.~\ref{fig_visual_abstact}, the tiles enable coil-induced emission of electromagnetic fields (yellow) that interact with the mover's field (blue). During operation, the movers hover above the tiles and can be controlled in six dimensions by adjusting the coil's currents contained in the tiles. The six degrees of freedom (DoFs) of a mover are visualized in Fig.~\ref{fig_mover_dofs}. MagLev systems are already included in the first industrial applications. The most advanced systems available are XBot (Planar Motor), ACOPOS 6D (B\&R), XPlanar (Beckhoff Automation), and ctrlX $\text{FLOW}^{\text{6D}}$ (Bosch Rexroth). Industrial applications with MagLev systems typically focus on product transport and do not consider the manipulation of objects. However, to increase productivity and exploit the full potential of MagLev systems, material flow and manipulation should be combined in the future. Therefore, MagLev systems can be seen as a special kind of robot. Hence, we introduce the more general term \emph{Magnetic Robotics}.

Developing and evaluating intelligent motion planning algorithms for MagLev systems requires an adequate physics-based simulation for several reasons. Firstly, MagLev systems are highly scalable, reconfigurable, and flexible, as the number of movers and tiles is only limited by computational resources, and the tiles can be assembled in almost any arbitrary planar layout. Therefore, the movers can be seen as a coordinated swarm or multi-agent system. In this research field, it is of high importance that motion planning algorithms allow valid high-quality results for a large number of movers in real time, since the movers are extremely agile. 
In this context, simulations can help to shift computational load into the offline phase reducing computation time during motion. Secondly, reinforcement learning (RL) is frequently used in multi-agent path planning~\cite{damani_primal_2_2021,chen_multi-agent_2022,guan_ab-mapper_2022}, and object manipulation~\cite{iannotta_can_2025,stranghoner_share-rl_2025,bergmann_precision-focused_2024,dengler_learning_2024,niemann_learning_2024,zeng_learning_2018}, where simulations are typically employed to accelerate training and evaluation.
Some MagLev systems also provide proprietary simulations, which are particularly designed for the respective system. These do not focus on algorithm development, and especially not on RL. Thus, currently, there is no physics-based system-agnostic simulation for MagLev systems available, which is designed for the development of motion planning and manipulation algorithms. In addition, benchmarks are missing to adequately compare new motion planning approaches. In this paper, we therefore present MagBotSim (\emph{Magnetic Robotics} Simulation), a physics-based simulation and RL environments for motion planning and object manipulation in the field of \emph{Magnetic Robotics}. We show that control policies trained using MagBotSim can be transferred to real MagLev systems, and we propose several benchmarks to make motion planning approaches for MagLev systems comparable. Moreover, we suggest future research directions for the high-level control of MagLev systems. MagBotSim is open-source, published under the GNU General Public License v3.0, and includes the following features:
\begin{itemize}
    \item Compatibility with common RL libraries.
    \item Basic environments without RL API, as well as with single-agent (Gymnasium~\cite{towers_gymnasium_2024}) and multi-agent (PettingZoo~\cite{terry_pettingzoo_2021}) RL APIs, that serve as starting points for easy development of new research-specific environments.
    \item \emph{Magnetic Robotics} utilities, e.g. impedance control for the movers and auxiliary functions to simplify measurements for the proposed benchmarks.
    \item Example environments for motion planning with MagLev systems, as well as tutorials on how to build new research-specific environments.
    \item Installation via PIP.
\end{itemize}
Therefore, MagBotSim is a unique starting point for the development, application, and evaluation of intelligent motion planning and manipulation capabilities that transform Magnetic Levitation to \emph{Magnetic Robotics}.
\section{Related Work}
\label{sec_related_work}
\textbf{Magnetic Levitation.} In medical applications, magnetic microrobots are frequently studied~\cite{isitman_trajectory_2025,sallam_autonomous_2024,song_motion_2023}. However, the MagLev systems we are considering are on a different scale and are typically used in industrial applications rather than medical setups, which require completely different motion planning approaches. In general, we distinguish between high-level control, i.e. motion planning and object manipulation, and low-level control, i.e. adjusting the magnetic fields. Recent studies focus on the development of appropriate hardware and low-level control~\cite{lu_6d_2012,trakarnchaiyo_design_2023,hartmann_end--end_2025}.
In terms of high-level control, Pierer von Esch et al.~\cite{pierer_von_esch_sensitivity-based_2025} evaluate a sensitivity-based distributed nonlinear model predictive (DMPC) scheme in four multi-agent scenarios using a MagLev system. Janning et al.~\cite{janning_conflict-based_2025} use conflict-based search together with model predictive control for multi-agent path finding. However, the authors report that their approach is limited to small numbers of movers. Nilsson, Ternerot, and Rehme~\cite{nilsson_henry_and_ternerot_johan_path_2022,rehme_louise_path-planning_2025} developed a multi-agent path-finding approach to line up movers on a prespecified path without collisions after a sudden failure of the system, while Ranc~\cite{ranc_path_2022} combined a path-finding algorithm with a deadlock-solving approach for multi-agent coordination. However, these approaches focus on transportation rather than manipulation. In general, the current literature lacks research on high-speed motion planning algorithms for MagLev systems, and no works deal with the manipulation of objects. In Section~\ref{sec_future_research}, we therefore suggest future research directions for MagLev systems.\\
\textbf{Simulation in Swarm Robotics.} Calder\'on-Arce et al.~\cite{calderon-arce_swarm_2022} provide a comprehensive overview of swarm robotics simulators, comparing platforms such as Stage~\cite{vaughan_massively_2008}, Webots~\cite{michel_cyberbotics_2004}, and ARGoS~\cite{pinciroli_argos_2012}. Since then, several new simulators have emerged, including MVSIM~\cite{blanco-claraco_multivehicle_2023} for ground vehicle fleets, SCRIMMAGE~\cite{demarco_simulating_2019} for large-scale UAV/UGV swarms, and py-bullet-drones~\cite{panerati_learning_2021} for drone swarms. These simulators are primarily designed for coordination tasks like exploration and foraging, with ground-based platforms assuming slow dynamics and uncertain localization. While drone simulators like SCRIMMAGE and py-bullet-drones also handle high-speed dynamics, they focus on aerial navigation and coordination. In contrast, MagBotSim focuses on industrial applications and targets the unique characteristics of MagLev systems.\\
\textbf{Reinforcement Learning Environments.} Several works in the field of RL and robotics have proposed simulation environments for different purposes: Gymnasium~\cite{towers_gymnasium_2024} and PettingZoo~\cite{terry_pettingzoo_2021}, which define standard APIs for single-agent and multi-agent RL, contain a collection of RL environments that serve as examples for algorithm tests and benchmarks. Gymnasium contains some simple robotics environments, e.g. for object pushing, but robotics is not the main focus.
The Minigrid and Miniworld libraries~\cite{chevalier-boisvert_minigrid_2023} contain 2D and 3D maze environments for RL research. The 2D Minigrid environments are simple grid worlds that are closely related to MagLev systems, since the tile layout can be understood as a grid. However, Minigrid is not based on a physics engine. Safety-Gymnasium~\cite{ji_safety_2023} and safe-control-gym~\cite{yuan_safe-control-gym_2022} focus on the safety of learned controllers, especially RL controllers, but it is not shown that trained policies are transferable to real robots. Isaac Gym~\cite{makoviychuk_isaac_2021} is a collection of high-performance robotics environments. The main focus of this work is a GPU-accelerated training pipeline. panda-gym~\cite{gallouedec_panda-gym_2021} contains a collection of RL environments for the Franka Emika Panda robot based on the PyBullet physics engine~\cite{coumans_pybullet_2016}. Similarly, Gymnasium-Robotics is a collection of RL robotics environments, but it is based on the MuJoCo physics engine~\cite{todorov_mujoco_2012} and the environments were introduced in different works, such as the OpenAI Fetch environments~\cite{plappert_multi-goal_2018}, the Franka Kitchen environments~\cite{gupta_relay_2020,fu_d4rl_2021}, various environments with a Shadow Dexterous Hand~\cite{melnik_using_2021,plappert_multi-goal_2018,rajeswaran_learning_2018}, multi-agent environments (MaMuJoCo)~\cite{peng_facmac_2021}, and maze environments~\cite{fu_d4rl_2021}. MagBotSim is partly inspired by panda-gym and especially Gymnasium-Robotics, and we hope that it will serve as a comparable library for \emph{Magnetic Robotics} in the future.
\section{Design Decisions}
\label{sec_design_dec}
\textbf{Physics Engine.} Since object manipulation is of great importance for \emph{Magnetic Robotics}, MagBotSim must be based on a physics engine. We chose the MuJoCo physics engine~\cite{todorov_mujoco_2012} for several reasons. MuJoCo is well-known in the robotics community, easy to install, and has a detailed documentation. In addition, MuJoCo has a very flexible actuator model that facilitates simple controller design. Therefore, it is possible to use various types of controllers, such as admittance or impedance controllers, which is important to enable the user to simulate a large variety of industrial applications.\\
\textbf{Simulation of Magnetic Fields.} A characteristic component of a MagLev system is the control of the movers with magnetic fields. Simulating interacting magnetic fields is challenging and computationally costly. Since RL is known for long training runs, our main focus is on an efficient simulation of MagLev systems for high-level control. Therefore, we currently do not simulate the magnetic fields, as this is not strictly necessary for motion planning or object manipulation. However, ignoring magnetic fields causes discrepancies between simulated and real mover behavior. In the real system, a low-level controller adjusts the magnetic fields, whereas in simulation, we currently use an impedance controller as a more efficient alternative. To support low-level controller development, MagBotSim will be extended with a magnetic field model (see Section~\ref{sec_future_research}).\\
\textbf{Collision Detection.} Motion planning approaches for multiple movers must avoid collisions and keep all movers within the system's boundaries, i.e. above a tile.
A typical strategy to ensure safety, even with position lags or noisy sensor data, is to add a safety margin to the actual size of a mover. Such properties are not originally supported by MuJoCo's collision detection. Instead, MagBotSim provides the possibility to check for collisions with additional safety margins.\\
\textbf{Reinforcement Learning API.} MagBotSim's RL environments follow either the Gymnasium~\cite{towers_gymnasium_2024} API (formerly OpenAI Gym~\cite{brockman_openai_2016}) for single-agent RL or the PettingZoo~\cite{terry_pettingzoo_2021} API for multi-agent RL. Both are well-known in the RL community and provide a standardized structure for RL environments. This allows MagBotSim to be used with common RL libraries, such as Stable-Baselines3~\cite{raffin_stable-baselines3_2021}.\\
\textbf{Custom Environments.} Another important design criterion was to provide the user with the ability to develop custom environments. In this context, the term ``environment'' does not only refer to RL environments, but to (industrial) environments in general, since MagBotSim can be used without the RL API.  MagBotSim generates a MuJoCo XML string to define the MuJoCo model.
\begin{wrapfigure}{r}{0.5\textwidth}
  \centering
  \vspace{-0.6cm}
  \includegraphics[width=0.5\textwidth]{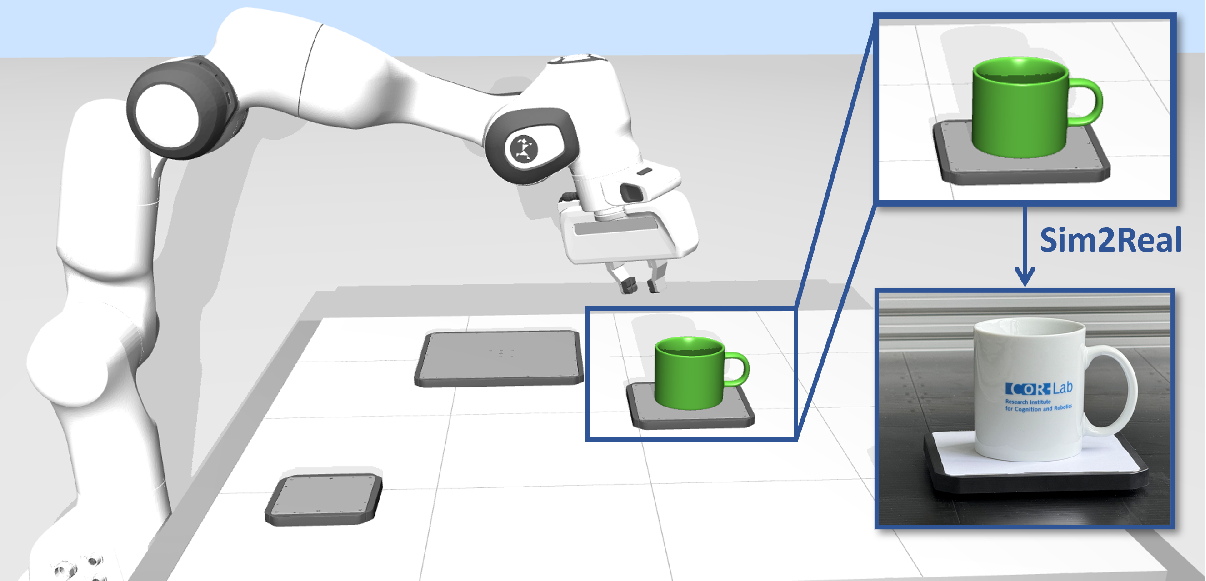}
  \caption{MagBotSim is designed to create customized environments that match real-world applications.}
  \vspace{-0.8cm}
  \label{fig_example_customizability}
\end{wrapfigure}
We decided to use XML strings instead of XML files, as loading an XML file is slower than generating an XML string. Since a reset of an (RL) environment might require building and compiling a new model to ensure that the simulation is physically correct, e.g. if the size of an object is changed, a new XML model might be generated frequently. Therefore, the generation process must be as fast as possible. MagBotSim automatically generates a MuJoCo XML string for a user-defined mover-tile configuration, which can be customized by defining additional XML strings, e.g. user-defined actuators for the movers, sensors, objects, or robots. In this way, the user can create a simulation of a specific (industrial) application. To show that it is possible to customize environments by adding robots and objects, an example of a custom environment with a Franka Emika Panda robot provided by MagBotSim is shown in Fig.~\ref{fig_example_customizability}. We also include a picture of a real mover with a mug to emphasize that MagBotSim is designed to match real-world applications. We provide examples and tutorials on how to use MagBotSim in the documentation.

\section{Benchmarks}
\label{sec_benchmarks}
\begin{figure}[h!]
    \vspace{-0.6cm}
    \centering
    \includegraphics[width=0.95\textwidth]{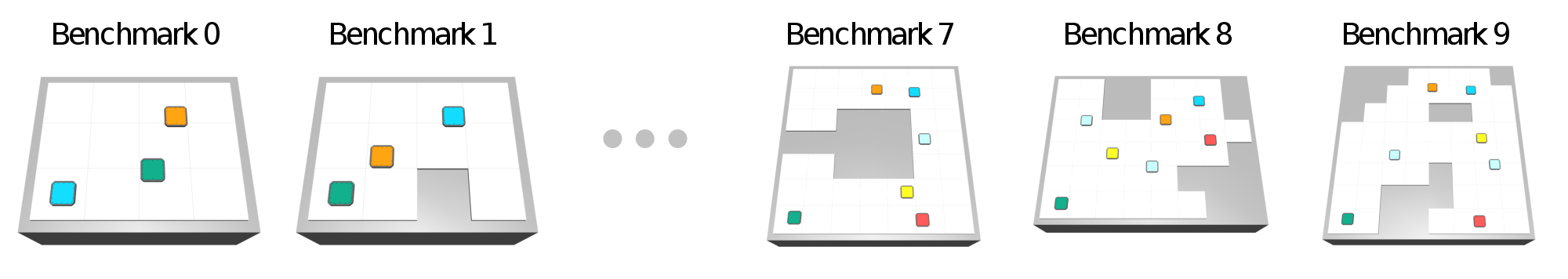}
    \caption{Benchmark environments included in MagBotSim.}
    \label{fig_benchmark_envs}
    \vspace{-0.4cm}
\end{figure}

To ensure the comparability of future research, we propose several task-specific benchmarks that allow users to measure the performance of suggested motion planning algorithms. We focus on object pushing and trajectory planning, since these are the tasks for which examples are included in MagBotSim, and we also provide the benchmark environments shown in Fig.~\ref{fig_benchmark_envs}. However, we will add new environments in the future and update the benchmarks accordingly. To simplify measurements for users, MagBotSim includes functions to measure specific properties, e.g. corrective movements. In the following, $n_1,n_2,n_3,n_4\in\mathbb{N}$ denote the total number of goals or the number of successfully reached goals.\\
\textbf{Benchmarks Object Pushing.} We propose the benchmarks listed in Table~\ref{tab_benchmarks_object_pushing} for object pushing tasks in which one object has to be pushed to a specific goal. The goal can either be a position or a pose (position and orientation). We distinguish two types of corrective movements, as suggested in our previous work~\cite{bergmann_precision-focused_2024}: overshoot and distance corrections.
\begin{table}
\centering
\scriptsize
\vspace{-0.3cm}
\caption{Benchmarks Object Pushing}\label{tab_benchmarks_object_pushing}
\begin{tabular}{|l|l|}
\hline
\textbf{Metric} &  \textbf{Measurement} \\
\hline
Success Rate &  Number of successfully reached goals divided by the total number of\\
             &  goals $n_1$.\\
\hline
Throughput &  Number of successfully reached goals $n_2$ divided by the time required to\\
           &  reach these goals.\\
\hline
Overshoot Corrections & Mean number of overshoot corrections for successfully reached goals $n_2$.\\
\hline
Distance Corrections & Mean number of distance corrections for successfully reached goals $n_2$.\\
\hline
           & Mean number of collisions for the total number of goals $n_1$ (including any\\
           & obstacle - object collisions). Of these:\\
Collisions & \textbf{Mover-Mover Collisions:} Mean number of mover-mover collisions if\\
           & more than one mover is used. \\
           & \textbf{Mover-Obstacle Collisions:} Mean number of collisions between a\\
           & mover and any static obstacle.\\
\hline
\end{tabular}
\end{table}\\
\textbf{Benchmarks Trajectory Planning.} We propose the benchmarks listed in Table~\ref{tab_benchmarks_trajectory_planning} for trajectory planning tasks in which one or typically many movers have to reach varying goals without collisions, where a mover-specific goal can be a position or a pose.
\begin{table}
\centering
\scriptsize
\vspace{-0.3cm}
\caption{Benchmarks Trajectory Planning}\label{tab_benchmarks_trajectory_planning}
\begin{tabular}{|l|l|}
\hline
\textbf{Metric} &  \textbf{Measurement} \\
\hline
Success Rate &  Number of successfully reached goals divided by the total number of goals $n_1$.\\
\hline
Makespan & Number of milliseconds required by the slowest mover to reach the total number\\
         & of goals $n_2$.\\
\hline
Throughput &  Number of milliseconds required by all movers to reach the total number of\\
           &  goals $n_3$. \\
\hline
           & Number of collisions for the total number of goals $n_1$. Of these:\\
           & \textbf{Mover-Mover Collisions:} Number of mover-mover collisions if more\\
Collisions & than one mover is used. \\
           & \textbf{Mover-Obstacle Collisions:} Number of collisions between a mover\\
           & and any static obstacle.\\
\hline
Smoothness & Mean weighted sum of jerk, acceleration, and velocity of all movers within the\\
           & total time period $t\in\mathbb{N}$.\\
\hline
Process Time & Process time required by all movers to reach the total number of goals $n_4$.\\
\hline
\end{tabular}
\vspace{-0.8cm}
\end{table}
\section{Experiments}
\subsection{Scalability}
\begin{figure}
  \centering
  \includegraphics[width=0.76\linewidth]{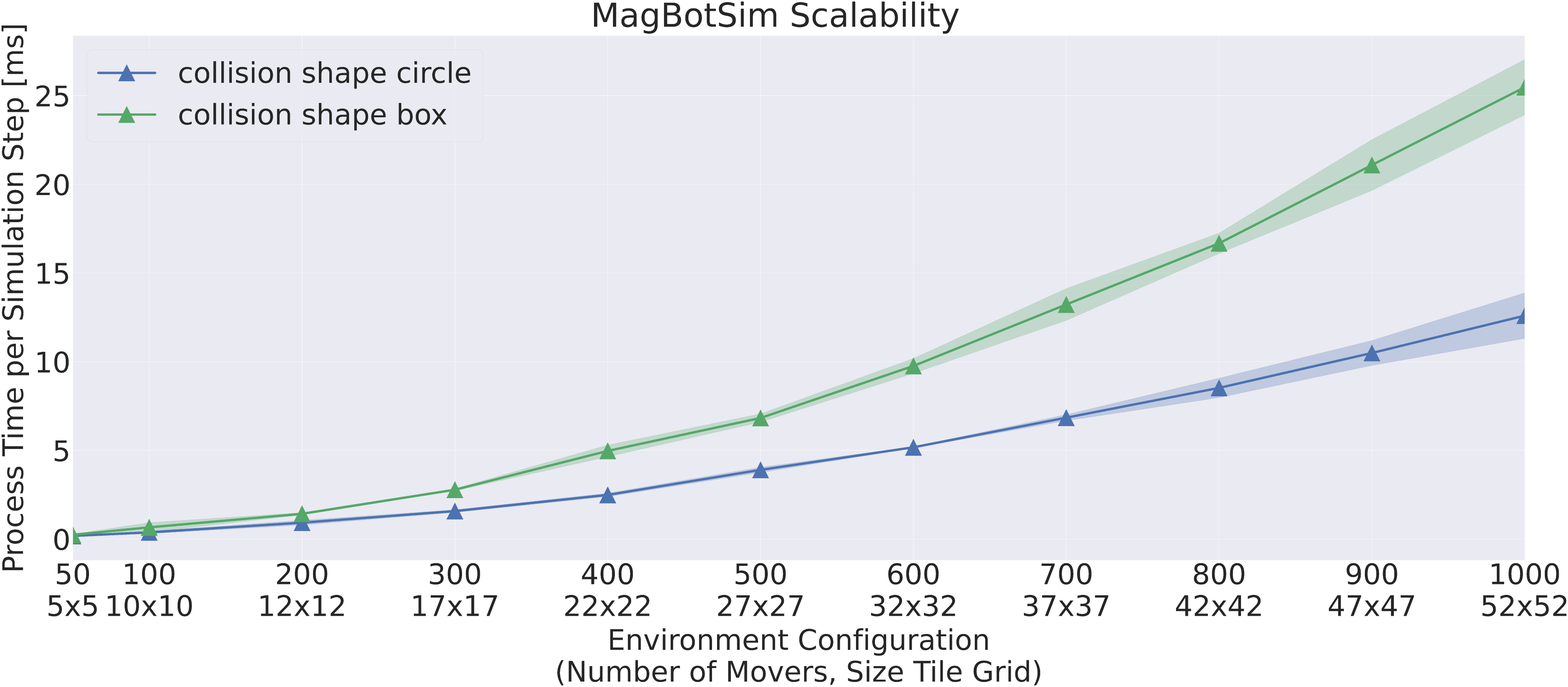}
  \caption{On a laptop CPU, it takes about $25\,$ms to execute one simulation step (apply controls, one MuJoCo integrator step, and collision checking) with 1k movers.}
  \label{plot_scalability}
\end{figure}

In research fields related to swarm robotics and multi-agent systems, it is important to show that motion planning algorithms can quickly produce valid results for a large number of movers. To show that MagBotSim enables users to perform such experiments, we run a scalability test on a MacBook Pro (2020) with an M1 processor, $16\,$GB RAM, and macOS 15.6.1. We measure the process time that is required to execute one simulation step, which includes applying controls to all movers, one MuJoCo integrator step, and collision checking. The tiles are arranged in a grid with the same number of tiles in x and y directions without missing tiles within the grid. Each mover is placed in the center of a tile. Since the collision checking differs depending on the collision shape that is used to approximate a mover, we perform the scalability test for each collision shape that is currently supported by MagBotSim. The mean process time and standard deviations depending on the number of movers and the collision shapes, calculated from $100$ simulation steps for each mover-tile configuration, are shown in Fig. \ref{plot_scalability}. 
The process times scale quadratically for the box and circular collision shapes. One simulation step for 1k movers on a $(57,57)$ tile grid takes about $25\,$ms when the box collision shape is used, and about $12\,$ms in the case of the circular collision shape. Thus, MagBotSim can be used with a large number of movers, even on a laptop.
\subsection{Object Manipulation in Simulation}
\begin{figure}
    \centering
    \includegraphics[width=0.85\textwidth]{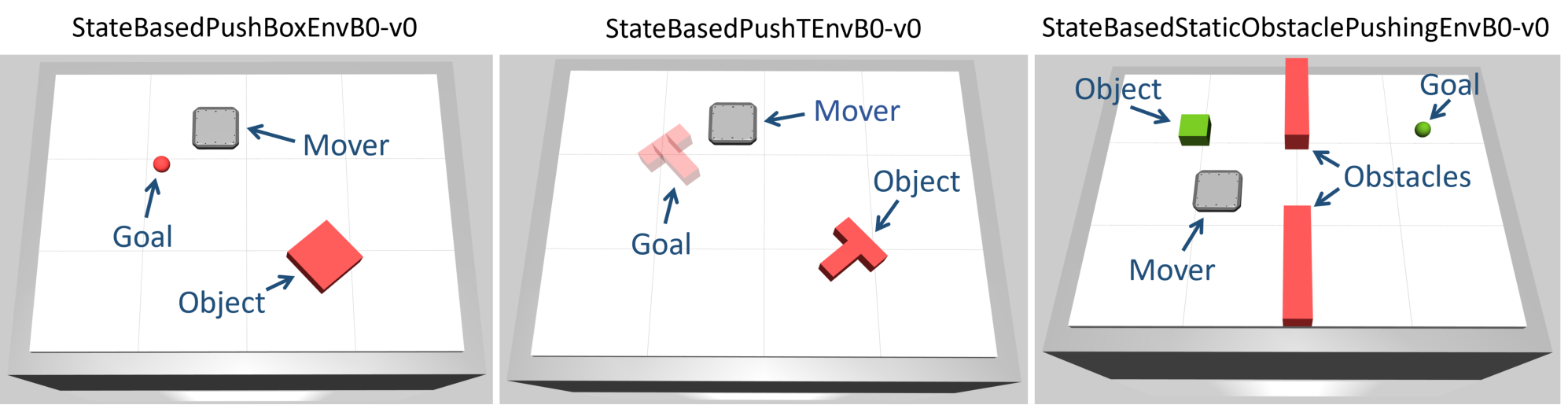}
    \caption{MagBotSim includes object pushing environments with and without obstacles.}
    \label{fig_pushing_envs}
\end{figure}

We consider pushing tasks, in which one mover has to push an object from various start positions to different goal positions, as examples for object manipulation with MagLev systems. Our pushing environments are configurable, i.e. object properties, such as shape, mass, and friction, can be changed. For example, we support T-shaped, L-shaped, and X-shaped objects, as well as boxes and cylinders. However, in the tasks considered in our experiments, the object's properties remain unchanged, but the start positions of the mover are chosen randomly. We consider three different tasks:
\begin{itemize}
    \item \texttt{Push-Box}. A box must be pushed to a goal position without the mover leaving the tiles. The goal is reached if the Euclidean distance between the object and the target is less than or equal to $5\,cm$.
    \item \texttt{Push-T}. A T-shaped object must be pushed to a goal pose without the mover leaving the tiles. The goal is reached if the coverage between the object and the target is greater than or equal to $0.7$.
    \item \texttt{PushWithObstacles}. A box must be pushed to a goal position without the mover leaving the tiles or the mover or the object colliding with one of the obstacles. The goal is reached if the Euclidean distance between the object and the target is less than or equal to $5\,cm$.
\end{itemize}
The corresponding MagBotSim environments are shown in Fig.~\ref{fig_pushing_envs}.\\The \texttt{StateBasedPushBoxEnvB0} is the MagLev version of the \texttt{FetchPush} environment~\cite{plappert_multi-goal_2018}. Both the \texttt{Push-Box} and the \texttt{Push-T} tasks are typical robotic benchmark tasks~\cite{andrychowicz_hindsight_2017,chi_diffusion_2023}. We train RL agents\footnote{All training and evaluation parameters are listed in our Git repository: \url{https://github.com/ubi-coro/MagBotSim/tree/main/docs/parameters}} to solve the pushing tasks using Soft Actor-Critic (SAC)~\cite{haarnoja_soft_2018} and Hindsight Experience Replay (HER)~\cite{andrychowicz_hindsight_2017}. The agents control the mover by specifying accelerations in x- and y-directions, as the low-level controller of a MagLev system requires smooth trajectories. The remaining DoFs of the mover are controlled by an impedance controller to keep the mover's orientation and hovering height. The benchmark measurements defined in Chapter~\ref{sec_benchmarks} are repeated $100\,$times. Our results are shown in Table~\ref{tab_results_object_pushing}. All agents learned to solve their pushing task, achieving success rates up to $99.56\%$ while almost completely avoiding collisions. The \texttt{Push-T} agent performs the most corrective movements, since the task requires adjusting not only the position, but also the orientation of the object. These results show that MagBotSim can be used to develop new object manipulation approaches for MagLev systems. 

\begin{table}
\centering
\scriptsize
\caption{Results Object Pushing}\label{tab_results_object_pushing}
\begin{tabular}{|l|c|c|c|c|}
\hline
\textbf{Metric} & \textbf{$n_1,n_2$} &  \textbf{Push-Box} & \textbf{Push-T} & \textbf{PushWithObstacles} \\
\hline
Success Rate ($\%$) & $200$ & $99.56\pm0.44$ & $73.13\pm2.96$ & $91.26\pm1.92$\\
\hline
Throughput (goals/s) & $100$ & $1.1829\pm0.0331$ & $0.5099\pm0.0172$ & $0.9665\pm0.0322$\\
\hline
Overshoot Corrections & $100$ & $0.0061\pm0.0081$ & $0.3241\pm0.0546$ & $0.0036\pm0.0061$\\
\hline
Distance Corrections & $100$ & $0.2822\pm0.0611$ & $2.0763\pm0.119$ & $0.1346\pm0.0394$\\
\hline
Collisions & $200$ & $0.002\pm0.0033$ & $0.0046\pm0.0041$ & $0.0451\pm0.0129$\\
\hline
Mover-Mover Collisions & $200$ & --- & --- & ---\\
\hline
Mover-Obstacle Collisions & $200$ & $0.002\pm0.0033$ & $0.0046\pm0.0041$ & $0.007\pm0.006$\\
\hline
\end{tabular}
\end{table}
\subsection{Sim2Real}
\begin{wrapfigure}{r}{0.48\textwidth}
    \centering
    \vspace{-1.05cm}
    \includegraphics[width=0.48\textwidth]{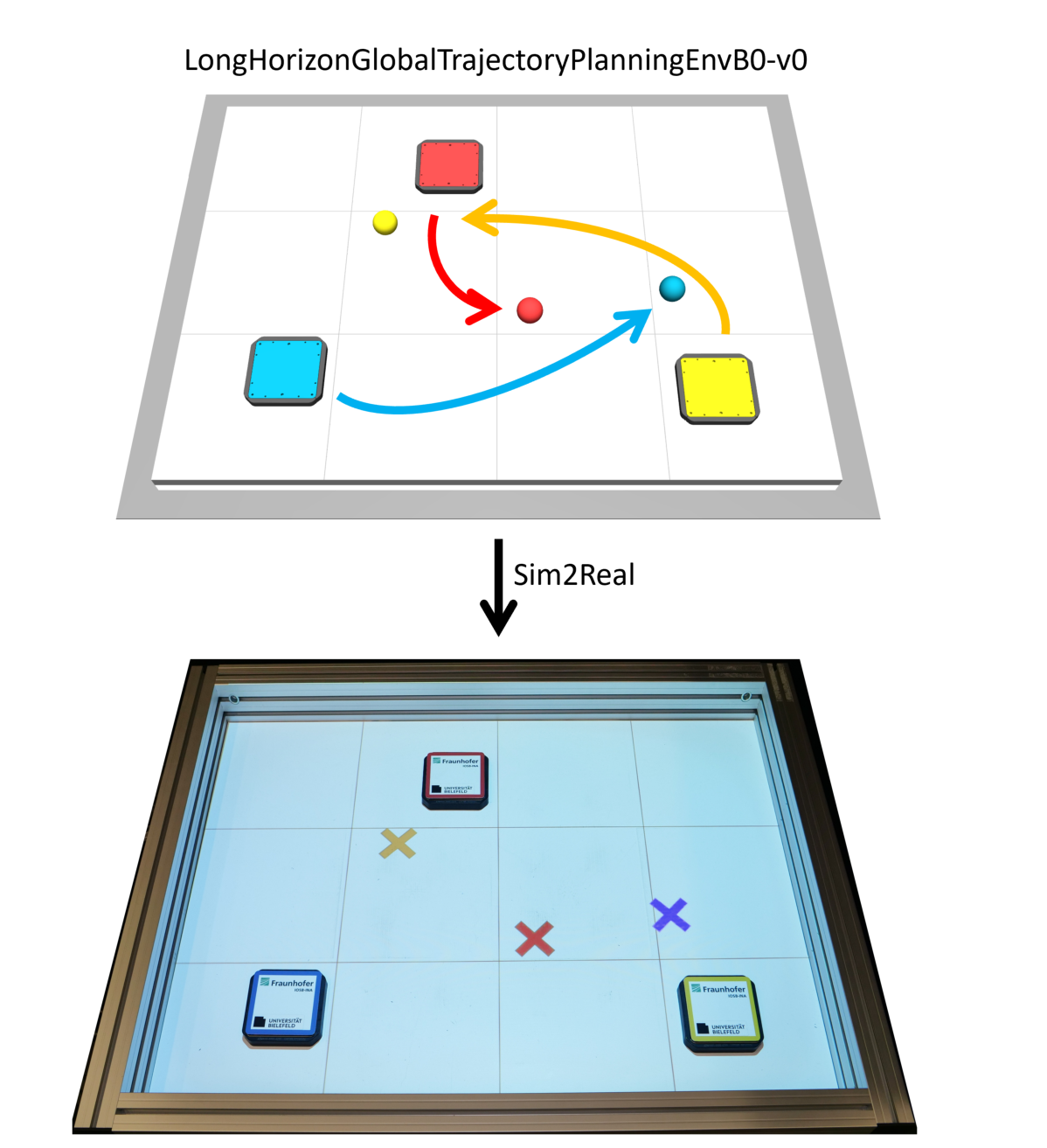}
    \caption{A trajectory planning agent trained with MagBotSim can be transferred to a real MagLev system, e.g. the XPlanar system (Beckhoff Automation).}
    \label{fig_trajectory_planning_sim2real}
    \vspace{-0.7cm}
\end{wrapfigure}
To show that control policies can be transferred to a real MagLev system, we train an RL agent to plan trajectories for three movers on a $4x3$ tile grid from various start positions to different goal positions without collisions. A collision is detected if any two movers collide or if a mover leaves the tile surface. To train the agent, we use TorchRL's~\cite{bou_torchrl_2023} SAC and HER implementation. A mover reaches its goal if the Euclidean distance between the mover and the goal is less than or equal to $10\,$cm. As shown in Fig.~\ref{fig_trajectory_planning_sim2real}, we trained the agent with MagBotSim and transferred it to a Beckhoff XPlanar system. The benchmark measurements defined in Chapter~\ref{sec_benchmarks} are repeated $100\,$times in simulation and $3\,$times on a real system. Our results are summarized in Table~\ref{tab_sim2real_results}. The agent achieves success rates close to $100\%$ on the real system and makespans and throughputs that are close to the ones measured in simulation. Thus, our results demonstrate that policies trained with MagBotSim can be executed on a real system without additional training or calibration. The TwinCAT3 project used for the sim2real transfer is publicly available on our website and serves as an example for using a control policy trained with MagBotSim in real time to control the movers of a Beckhoff XPlanar system.
\begin{table}
\centering
\scriptsize
\caption{Results Trajectory Planning}\label{tab_sim2real_results}
\begin{tabular}{|l|c|c|c|}
\hline
\textbf{Metric} & \textbf{$n_1,n_2,n_3,n_4,t$} &  \textbf{Simulation} & \textbf{Real System} \\
\hline
Success Rate ($\%$) & $15000$ &  $99.3\pm0.001$ & $99.97\pm0.02$\\
\hline
Makespan (ms) & $1000$ & $511463.58\pm11108.03$ & $530970.33\pm14461.78$\\
\hline
Throughput (ms) & $3000$ & $495261.89\pm7345.36$ & $502553.00\pm6412.13$ \\
\hline
Collisions & $15000$ & $32.09\pm8.49$ & $0.0\pm0.0$ \\
\hline
Mover-Mover Collisions & $15000$ & $26.45\pm7.66$ & $0.0\pm0.0$\\
\hline
Mover-Obstacle Collisions & $15000$ & $5.64\pm2.01$ & $0.0\pm0.0$\\
\hline
Smoothness & $1000\,$ms & $4610.97\pm490.87$ & --- \\
\hline
Process Time & $300$ & $0.1178\pm0.0035$ & --- \\
\hline
\end{tabular}
\end{table}
\section{Future Work and Research Directions}
\label{sec_future_research}
\textbf{Hardware Acceleration.} Currently, MagBotSim does not support hardware-accelerated environments. Therefore, we plan to investigate whether a hardware-accelerated version of MagBotSim, e.g. using JAX~\cite{bradbury_jax_2018}, significantly improves the performance compared to the CPU-only version.\\
\textbf{Magnetic Field Simulation.} MagBotSim currently does not contain a magnetic field simulation. However, in order to extend MagBotSim to also include low-level control environments, a magnetic field simulation is required. Therefore, future research should focus on developing an efficient, hardware-accelerated magnetic field simulation, as well as low-level control environments, e.g. to plan energy optimized motions or to eliminate vibrations generated by the low-level controller depending on the current payload of a mover.\\
\textbf{Sloshing-Free Movements.} In current research~\cite{heins_keep_2023,muchacho_solution_2022,moriello_manipulating_2018}, non-prehensile object manipulation and anti-sloshing are frequently discussed. Note that the mug, shown in Fig. \ref{fig_example_customizability}, is not attached to the mover. This requires that motion planning approaches generate sloshing-free movements to prevent the mug from falling off the mover or spilling contained liquids, which is a non-prehensile object manipulation task that should be investigated in the future.\\
\textbf{High-Speed Motion Planning Approaches and Safety.} Motion planning approaches for \emph{Magnetic Robotics} have to meet strong requirements due to the flexibility of the system and the agility of the movers. More specifically, the development of high-speed planning algorithms is demanding as the state of the entire system can change completely within fractions of a second. The XPlanar system from Beckhoff Automation achieves a maximum velocity of $2\,$m/s and a maximum acceleration of $10\,$m/s². The low-level controller operates with a cycle time of $250\,\mu$s, whereas a high-level controller uses a cycle time of $1\,$ms. This is one of the two main differences compared to mobile robots, which typically move comparatively slowly. The second difference is that, in a MagLev system, the positions of all movers are precisely known at each time step. Therefore, global motion planning algorithms for \emph{Magnetic Robotics} can also be considered. In current industrial applications,  movers usually follow manually engineered and predefined paths. This considerably simplifies the complexity of the motion planning problem, but also reduces the flexibility of the system. To avoid sub-optimal solutions by using the full flexibility, we consider RL to be suitable for the development of intelligent motion planning approaches for several reasons: RL
provides the possibility to learn optimal sequences of actions. In addition, deep RL allows the use of neural networks, which are known for their generalization capability, while providing the ability to work with high-dimensional inputs and outputs, as well as a deterministic inference time, which makes them particularly suitable for high-speed computations and real time requirements. In addition, the motion planning approaches must also provide mathematically provable safety guarantees to be certified for use in industrial applications. As MagBotSim's environments are compatible with single-agent and multi-agent RL APIs, we hope it will serve as a starting point for the development of new high-speed, but safe, motion planning approaches.\\
\textbf{Collaborative Object Manipulation.} Motion planning for collaborative object manipulation tasks is complicated using robot arms due to safety requirements, i.e. collision avoidance. In contrast, the movers of a MagLev system can easily collaborate, e.g. to jointly push or carry objects that are too heavy for a single mover. Additionally, during transport, movers can manipulate objects, e.g. mixing liquids. Thus, combining product transport and object manipulation using MagLev systems increases productivity and is therefore more profitable for a company than relying solely on a MagLev system for transportation. Moreover, movers can be coupled, e.g. by a kinematic with a gripper that is controlled by the movements of the movers. More generally, these kinematics can be designed for specific tasks within a machine. This way, a machine becomes easily adjustable without the typically high development costs. Future research will focus on developing motion planning algorithms that can handle coupled movers and enable collaboration between movers while being both fast and safe. 
\section{Conclusion}
We introduced MagBotSim, a physics-based simulation and RL environments for object manipulation and trajectory planning with MagLev systems. We showed that control policies trained with MagBotSim can be transferred to a real MagLev system without any additional training or calibration. In addition, MagBotSim can run on a laptop even if 1k movers are used and allows building custom research-specific environments with and without an RL API. Moreover, we proposed several benchmarks to make future motion planning approaches for MagLev systems comparable. By suggesting future research directions for the high-level control of MagLev systems, we hope that this work is a starting point for the development of motion planning algorithms for \emph{Magnetic Robotics}. 
\\ \\
{\normalfont\small \textbf{Disclosure of Interests.} The authors have no competing interests to declare that are relevant to the content of this article.}
%
%
\bibliographystyle{splncs04}
\bibliography{ReferencesMagBotSim}
%
%
%
%
%
\end{document}